\documentclass{article} 
\usepackage{iclr2015,times}

\usepackage[hyphens]{url}
\usepackage{amsmath}
\usepackage{mathtools}
\usepackage{caption}
\usepackage{float}
\usepackage{xcolor}
\usepackage{graphicx}
\usepackage{listings}
\usepackage{amsfonts}
\usepackage{arydshln}

\DeclarePairedDelimiter\ceil{\lceil}{\rceil}
\DeclarePairedDelimiter\floor{\lfloor}{\rfloor}

\title{Fast Convolutional Nets With \fbfft: \\A GPU Performance Evaluation}

\author{
\textbf{Nicolas Vasilache, Jeff Johnson, Michael Mathieu,}\\
\textbf{Soumith Chintala, Serkan Piantino \& Yann LeCun} \\
Facebook AI Research \\
770 Broadway, New York, NY 10003, USA \\
\texttt{\{ntv,jhj,myrhev,soumith,spiantino,yann\}@fb.com}
}

\newcommand{\cnn}{CNN~}
\newcommand{\cnnend}{CNN}
\newcommand{\cnns}{CNNs~}
\newcommand{\cnnsend}{CNNs}
\newcommand{\cublas}{cuBLAS~}
\newcommand{\cublasend}{cuBLAS}
\newcommand{\cuda}{CUDA~}

\newcommand{\cudnn}{cuDNN~}
\newcommand{\cudnnend}{cuDNN}
\newcommand{\cufft}{cuFFT~}
\newcommand{\cufftend}{cuFFT}
\newcommand{\dft}{DFT~}

\newcommand{\fbfft}{{\tt fbfft}~}
\newcommand{\fbfftend}{{\tt fbfft}}
\newcommand{\fbcuda}{{\tt fbcuda}~}

\newcommand{\fbcunn}{{\tt fbcunn}~}

\newcommand{\gpu}{GPU~}
\newcommand{\gpuend}{GPU}
\newcommand{\gpus}{GPUs~}
\newcommand{\gpusend}{GPUs}
\newcommand{\nvidia}{NVIDIA~}
\newcommand{\nvidiaend}{NVIDIA}
\newcommand{\fft}{FFT~}
\newcommand{\fftend}{FFT}
\newcommand{\ffts}{FFTs~}
\newcommand{\fftsend}{FFTs}
\newcommand{\ifft}{IFFT~}
\newcommand{\ifftend}{IFFT}
\newcommand{\onedfft}{1-D~FFT~}
\newcommand{\onedfftend}{1-D~FFT}

\newcommand{\twodfft}{2-D~FFT~}
\newcommand{\twodfftend}{2-D~FFT}

\newcommand{\dif}{DIF~}
\newcommand{\difend}{DIF}
\newcommand{\dit}{DIT~}
\newcommand{\ditend}{DIT}

\iclrfinalcopy
\iclrconference

\begin{document}

\maketitle

\begin{abstract}
We examine the performance profile of Convolutional Neural Network (\cnnend)
training on the current generation of \nvidia Graphics Processing Units (\gpusend). We
introduce two new Fast Fourier Transform convolution implementations: one based on \nvidiaend's
\cufft library, and another based on a Facebook authored \fft implementation, \fbfftend,
that provides significant speedups over \cufft (over $1.5 \times$) for whole \cnnsend.
Both of these convolution implementations are available in open source,
and are faster than \nvidiaend's \cudnn implementation for many common convolutional layers
(up to $23.5 \times$ for a synthetic kernel configuration).
We discuss different performance regimes of convolutions, comparing areas where straightforward
time domain convolutions outperform Fourier frequency domain convolutions. Details on algorithmic
applications of \nvidia \gpu hardware specifics in the implementation of \fbfft are also provided.
\end{abstract}

\section{Introduction}
Deep convolutional neural networks (\cnnsend) have emerged as one of the most promising techniques
to tackle large scale learning problems, whether in image and face
recognition, audio and speech processing or natural language understanding.
A convolutional layer within these networks provides useful properties such as translation equivariance of activations.
A limiting factor for use of convolutional nets on large data sets was, until recently, their
computational expense.

\cite{NIPS2012_4824} demonstrated that training of large \cnns with millions of
weights and massive data sets is tractable when graphics processing units (\gpusend) are properly put to use.
Since then, renewed interest in \cnns insufflated a fresh breath in various frameworks
and implementations, including Torch (\cite{collobert:2011c}), Theano (\cite{bergstra+al:2010-scipy}),
cuda-convnet (\cite{cudaconv2}) and Caffe (\cite{jia2014caffe}).
Many of these frameworks are based around codes for \nvidia \gpus using \cuda (\cite{gpuoverview}).

We discuss our contributions to
convolution performance on these \gpusend, namely using Fast Fourier Transform (\fftend)
implementations within the Torch framework.
We summarize the theory behind training convolutional layers both in the time and frequency domain in
Section~\ref{sec:math}. We then detail our implementations.
The first is based on \nvidiaend's \cufft and \cublas libraries (Section~\ref{sec:cufft}).
We evaluate our relative performance to \nvidiaend's \cudnn library (\cite{cudnn})
on over $8,000$ different configurations (Section \ref{sec:perf}).
We significantly outperform \cudnn and other time domain convolution implementations for
a wide range of problem sizes.

Our second implementation is motivated by limitations in using a black box library
such as \cufft in our application domain, which we describe. In reaction, we implemented a from-scratch
open-source implementation of batched \onedfft and
batched \twodfftend, called Facebook \fft (\fbfftend), which achieves over $1.5\times$ speedup over \cufft for the
sizes of interest in our application domain. This implementation achieves \gpu
efficiency ratios of over $75\%$ in certain cases. We describe an ongoing
effort to further improve the performance of our solution
based on algorithmic tiling (Section \ref{sec:future}) before we
conclude. Our implementation is released as part of the \fbcuda and \fbcunn
open-source libraries at \url{http://github.com/facebook}.

\section{Convolution}
\label{sec:math}

Discrete convolution and cross-correlation are used in \cnnsend.
We quickly summarize these and their implementation,
with a formulation mirroring \cite{DBLP:journals/corr/MathieuHL13}.
Forward propagation (\emph{fprop}) inputs are a set $f$ of \emph{input feature planes} $x_i, i \in f$.
These are cross-correlated\footnote{Torch practice is that the forward pass is
cross-correlation, hence the $\star$.} with $f' \times f$ different
filter kernel weights $w_{(j, i)}, j \in f', i \in f$, producing \emph{output feature planes} $y_{j}, j \in f'$.
Each input and output feature can be part of a \emph{minibatch} $S$,
so we have $x_{(s, i)}$ and $y_{(s, j)}, i \in f, j \in f', s \in S$:

$$y_{(s, j)} = \sum_{i \in f} x_{(s, i)} \star w_{(j, i)}$$

The feature planes $f$ are \emph{reduced} (summed) pointwise.
For back-propagation (\emph{bprop}), the gradient of the loss with respect to outputs are convolved with the kernels:

$$\frac{\partial L}{\partial x_{(s, i)}} = \sum_{j \in f'} \frac{\partial L}{\partial y_{(s, j)}} * w_{(j, i)}$$

Reduction is over $f'$ here.
Finally, the kernel weights are updated using the gradient of the loss with respect to the weights (\emph{accGrad}):

$$\frac{\partial L}{\partial w_{(j, i)}} = \sum_{s \in S} \frac{\partial L}{\partial y_{(s, j)}} \star x_{(s, i)}$$

Reduction is over $S$ here. For purposes of this paper, we use set symbols
interchangeably to refer to their size: each input plane is a 2-D matrix
of size $h \times w$, and each filter kernel is a 2-D matrix of size $k_h \times k_w$\footnote{2-D can be extended to $n$-D, $n \geq 1$.}.
The output planes $y_{(s, i)}$ are of size $(h - k_h + 1) \times (w - k_w + 1)$, and implement \emph{valid}-only
convolution, as per MATLAB terminology.
\emph{Input zero padding} and \emph{input mirror padding} around the margins of
the input ($p_h, p_w$) can be optionally added.\footnote{Input size $(h + p_h)
  \times (w + p_w)$,
output size $(h + p_h - k_h + 1) \times (w + p_w - k_w + 1)$.}


A popular convolution implementation is to \emph{unroll} the data until the
computation is in the form of a large matrix multiplication
(\cite{cnnmatrix}).
This is the strategy followed by many implementors, since matrix multiplication
is a well-tuned linear algebra primitive available on virtually any platform.
While it is possible to provide instances of direct calculation that are faster than
matrix unrolling (\emph{e.g.}, for large $S$, \cite{cudaconv2}), it is challenging to provide an
implementation that is faster for more than just a small subset of possible convolution problems.

Introducing strides in this form of convolution
(\emph{i.e.}, performing the convolution at every $d_h, d_w$-th offset) is a popular way
to reduce the computational cost at the expense of precision.
The memory accesses required are very similar but with fewer reuse opportunities.
On the other hand, by the \emph{convolution theorem}, a convolution of two discrete signals can
be performed with lower asymptotic complexity by performing the multiplication
in the frequency domain.
Applied to the forward pass, it becomes:

$$y_{(s, j)} =  \sum_{i \in f} x_{(s, i)} \star w_{(j, i)} = \sum_{i \in f} \mathcal{F}^{-1}\left(\mathcal{F}(x_{(s, i)}) \circ \mathcal{F}(w_{(j, i)})^*\right)$$

where $^*$ denotes complex conjugation and $\circ$ is the pointwise product. The discrete Fourier basis used
is the largest of the two components convolved and the output.\footnote{$(h \times w)$-dimensional or even bigger for performance (Section \ref{sec:design_space}).}
Linearity of the \dft allows one to perform the sum above in the Fourier domain if desired.
Applying the \fft then yields a $\mathcal{O}(S f  f'  n^2 + (S f + f f' + S f') n^2 \log n)$
procedure in lieu of the original $\mathcal{O}(S  f  f'  n^2  k^2), n = h = w, k = k_h = k_w$.
Similar transformations apply for the other two passes.
We call this a \emph{frequency domain} convolution, in contrast to \emph{time domain} convolution via direct computation.
Strided convolutions via \fft can be implemented efficiently to obtain good
performance~\cite{DBLP:journals/neco/BroschT15}. We do not consider those in this paper.


\section{\cufft Convolution Implementation}
\label{sec:cufft}
In this section we discuss implementation strategies using the
\nvidia \cufft libraries and their efficiency.

\subsection{FFT Convolution Details}
\label{sec:computation_details}
We described the general formulation for the three types of convolutions in
section~\ref{sec:math}. Here, we borrow the Torch naming convention: \emph{input} for $x_{(s, i)}$;
\emph{weight} for $w_{(j, i)}$;
\emph{output} for $y_{(s, j)}$;
\emph{gradOutput} for $\partial L / \partial y_{(s, j)}$;
\emph{gradInput} for $\partial L / \partial x_{(s, i)}$;
and \emph{gradWeight} for $\partial L / \partial w_{(j, i)}$.
All are stored as single-precision floating point 4-D tensors in row-major layout, and are stored in
memory using the so-called \emph{BDHW} format. This is explicit in the expression
$In_{S \times f \times h \times w}$, with input image row data as the \emph{innermost} or
\emph{most varying} dimension.

Table~\ref{tab:passes} describes the in-order operations for FFT computation
of the forward pass, using the $FFT2D$ and $IFFT2D$ operators and {\tt Cgemm}
matrix multiplication. Similar implementations follow for the other two passes.
The \emph{G} prefix denotes gradients.
The \emph{F} suffix denotes $\mathbb{C}$-valued frequency domain tensors; the rest
are over $\mathbb{R}$.
The \emph{T} suffix denotes transposed tensors.

\begin{table}[h]
\caption{Implementation detail for forward propagation}
\label{tab:passes}
\begin{center}
\begin{tabular}{lll}
\multicolumn{1}{c}{\bf INPUT}  &\multicolumn{1}{c}{}  &\multicolumn{1}{c}{\bf OUTPUT}
\\ \hline \\
  $In_{S\times f \times h \times w}$ &
  $\xrightarrow{FFT2D}$ &
  $InF_{S\times f \times (h+p_h) \times (\floor{\frac{w+p_w}{2}}+1)}$ \\
  $Wei_{f' \times f\times k_h \times k_w}$ &
  $\xrightarrow{FFT2D}$ &
  $WeiF_{f'\times f \times (h+p_h) \times (\floor{\frac{w+p_w}{2}}+1)}$  \\
  $InF_{S\times f \times (h+p_h) \times (\floor{\frac{w+p_w}{2}}+1)}$ &
  $\xrightarrow{Trans2D}$ &
  $InFT_{(h+p_h) \times (\floor{\frac{w+p_w}{2}}+1)\times S\times f}$  \\
  $WeiF_{f' \times f\times (h+p_h) \times (\floor{\frac{w+p_w}{2}}+1)}$ &
  $\xrightarrow{Trans2D}$ &
  $WeiFT_{(h+p_h) \times (\floor{\frac{w+p_w}{2}}+1)\times f'\times f}$  \\
  $
  \left\{
  \begin{array}{c}
    InFT_{(h+p_h) \times (\floor{\frac{w+p_w}{2}}+1) \times S\times f}  \\
    WeiFT_{(h+p_h) \times (\floor{\frac{w+p_w}{2}}+1)\times f'\times f}^*
  \end{array}
  \right.
  $
  &
  $\xrightarrow{\tt Cgemm}$ &
  $OutFT_{(h+p_h) \times (\floor{\frac{w+p_w}{2}}+1) \times S \times f'}$  \\
  $OutFT_{(h+p_h) \times (\floor{\frac{w+p_w}{2}}+1) \times S \times f'}$ &
  $\xrightarrow{Trans2D}$ &
  $OutF_{S \times f' \times (h+p_h) \times (\floor{\frac{w+p_w}{2}}+1) }$ \\
  $OutF_{S \times f' \times (h+p_h) \times (\floor{\frac{w+p_w}{2}}+1)}$ &
  $\xrightarrow{IFFT2D}$ &
  $Out_{S \times f' \times (h-k_h+1) \times (w-k_w+1)}$ \\
\end{tabular}
\end{center}
\end{table}

Exact tensor dimensions are also given above. By taking advantage of the Hermitian
symmetry property of the 2-D DFT for $\mathbb{R}$-valued inputs we only store
about half the complex entries; the remaining can be obtained by complex
conjugation. This results in array sizes such as
$\floor{\frac{w+p_w}{2}}+1$.
We also perform interpolation by \emph{zero-padding}, which serves multiple purposes. First, it is
necessary to handle boundary conditions.\footnote{In this case, we typically
 have $p_h=\floor{\frac{k_h}{2}}$ and $p_w=\floor{\frac{k_w}{2}}$.} Second, it is
required to interpolate all operands over the same Fourier basis.\footnote{All tensors are
zero-padded to $(h+p_h) \times (w+p_w)$ before $FFT2D$.} Finally, padding has an
impact on the \fft algorithm used in practice, as well as on the floating point
operation count of non-\fft operations (Section~\ref{sec:design_space}).

Following the conversion into frequency domain, we perform
transpositions to prepare the tensors for {\tt Cgemm} matrix multiplication library calls.
The transposition converts the \emph{BDHW} layout into \emph{HWBD}.
The transposition is currently out-of-place and implemented using the {\tt Cgeam}
routine; we are also considering our own, in-place transposition routine.
{\tt Cgemm} library calls are performed on
transposed tensors in the frequency domain. Casting the operation as a {\tt Cgemm}
call allows us to benefit from the heavily tuned \cublas routine.
Eventually, we transpose the result back into the \emph{BDHW} format and perform a 2-D
inverse \fftend. At this point, the resulting real tensor, always $(h+p_h) \times (w+p_w)$, is
clipped to the appropriate final size: $(h-k_h+1) \times (w-k_w+1)$ for fprop,
$h \times w$ for bprop, $k_h \times k_w$ for accGrad.

\subsection{\cufft Design Space}
\label{sec:design_space}
We now discuss implementation aspects we explored.
Multiple factors influence the computational efficiency of FFTs:
transform size $n$, $n$'s prime factor decomposition,
and whether batched or iterated single transforms are
applied. In the deep learning domain, it is commonplace to deal with small sizes, $n \neq 2^k$.
If $n$ has undesirable properties,
efficiency can drop by an order of magnitude.\footnote{
\url{http://docs.nvidia.com/cuda/cufft/index.html#accuracy-and-performance}}

\cufft implements FFTs with the ubiquitous Cooley-Tukey
algorithm (\cite{cooleytukey})
which takes advantage of trigonometric
equalities to recursively decompose and reuse computations. This is further discussed in
the Supplement. Decomposition is built on specialized kernels of fixed
sizes which correspond to the prime factor
decomposition of $n$. \cufft implements specialized building blocks for radix sizes
$2, 3, 5, 7$, and for sizes $n$ where $4 | n$, it can use more efficient kernels exploiting
the conjugate symmetry property.
When $n$ does not admit a prime factor decomposition using those radices only, the
expensive Bluestein algorithm is used (\cite{bluestein}).
Because our results are used in the time domain, we can in fact zero-pad the image and kernel
to perform the FFT at any larger size that may be handled more efficiently.
Exploiting more efficient, larger sizes should be balanced against the extra cost introduced
in the subsequent transposition and matrix multiplication steps.
Table~\ref{tab:fft_perf}'s last case is one in which the best
tradeoff is not easily guessed.
\cufft also has batched mode optimizations when multiple FFTs
of the same size are being performed.

\subsection{\cublas Design Space}
The \cublas library also comes with different implementations for batched and
single operation modes. We had the choice between $3$ implementation options:
\begin{itemize}
\item for larger batches over small matrices, the {\tt cublasCgemmBatched} library
  call;
\item for smaller batches over larger matrices, multiple {\tt cublasCgemm} calls
  from the host;
\item for intermediate batch and matrix sizes, devices of compute
  capability 3.5 and higher support dynamic parallelism which allows \cuda kernels
  to launch other kernels. This can be beneficial for many launches over small
  matrices.
\end{itemize}
Note that the discussion above applies to multiplications after
transposition. So the matrix size is either $S\times f$,  $S\times f'$ or
$f\times f'$ and the number of such matrices is $h\times w$.
Vendor libraries are usually optimized for throughput and not latency, so we
expect it to be more efficient for larger sizes along critical
dimensions (\emph{i.e.}, image size for the batch case and $S\times f$,  $S\times f'$ or
$f\times f'$ for the multiple kernel case).
Due to build system limitations we were not able to experiment with the
dynamic parallelism strategy; we leave this for future work.

At the system level, we use \cuda streams and buffering of all \cuda resources and
intermediate buffers to remove synchronization points across convolutions.
We are mindful of memory consumption;
to address this we keep one single buffered copy of each type of tensor
involved. This behavior is tailored for a bulk synchronous
execution of layers on a GPU and is not adapted for multiple asynchronous
convolutions on the same \gpuend.
The buffers are automatically expanded as required and reused as much as possible.

\subsection{Autotuning}
We combine the above implementation with a simple autotuning strategy.
We devise a strategy selection mechanism that runs once for each problem size
and caches the fastest strategy out of a few dozen for later reuse.
The autotuning strategy explores different possible Fourier basis
sizes that can be decomposed in powers for which \cufft has an
efficient implementation. In other words, for an \fft dimension of size $n$, we
explore the sizes $i \in [n,~ 2 \lfloor \log_2 n \rfloor]$ where
$i=2^a 3^b 5^c 7^d$. When the input size is a power of $2$, the
search space is reduced to a single point.
In addition to Fourier basis sizes, we weigh in various \cublas calls and
asynchronous modes.


\section{\cufft Convolution Performance}
\label{sec:perf}

\subsection{Performance versus \cudnnend: 8,232 configurations}

We compare our \cufft convolution results against \nvidiaend's \cudnn 1.0 library (\cite{cudnn}),
which contains one of the fastest, general purpose convolution methods for the \gpuend,
using matrix unrolling.
It has decent performance for many problem sizes thanks to heavy autotuning of
\cublas codes for different problems. It is a strong baseline for this reason.

Image \cnns to date have for the most part used
square input images and filters, though rectangular filters are valid for other problems
(notably text \cnnsend, \cite{Collobert}). Thus, we restrict ourselves to a 5-D problem domain
$\{S, f, f', n (= h = w), k (= k_h = k_w)\}$.
Much of this space is not used in practice. Some
areas are perhaps over-emphasized (large $S$, small $k$) due to current engineering concerns.
We evaluate \cudnn vs \cufftend-based convolution for Table \ref{tab:problem_sizes_eval}'s
$8,232$ configurations.\footnote{Parameterized on output rather than input
size $h, w$ because the implied $h = y + k_h - 1, w = y + k_w - 1$ will be valid for any choice of $k_h, k_w$.}

\begin{table}[h]
\caption{Configuration elements evaluated}
\label{tab:problem_sizes_eval}
\begin{center}
\begin{tabular}{ll}
\multicolumn{1}{c}{\bf DIMENSION}  &\multicolumn{1}{c}{\bf SIZES EVALUATED}
\\ \hline \\
Minibatch size ($S$) & $1$, $16$, $64$, $128$ \\
Input filters ($f$) & $1$, $4$, $16$, $64$, $96$, $128$, $256$ \\
Output filters ($f'$) & $1$, $4$, $16$, $64$, $96$, $128$, $256$ \\
Kernel h/w ($k = k_h = k_w$)& $3$, $5$, $7$, $9$, $11$, $13$ \\
Output h/w ($y = h - k_h + 1 = w - k_w + 1$) & $1$, $2$, $4$, $8$, $16$, $32$, $64$ \\
\end{tabular}
\end{center}
\end{table}

\begin{figure}[H]

\vspace{-5.7cm}
\begin{minipage}{.5\textwidth}
  \hspace{-1cm}
  \includegraphics[width=10.5cm]{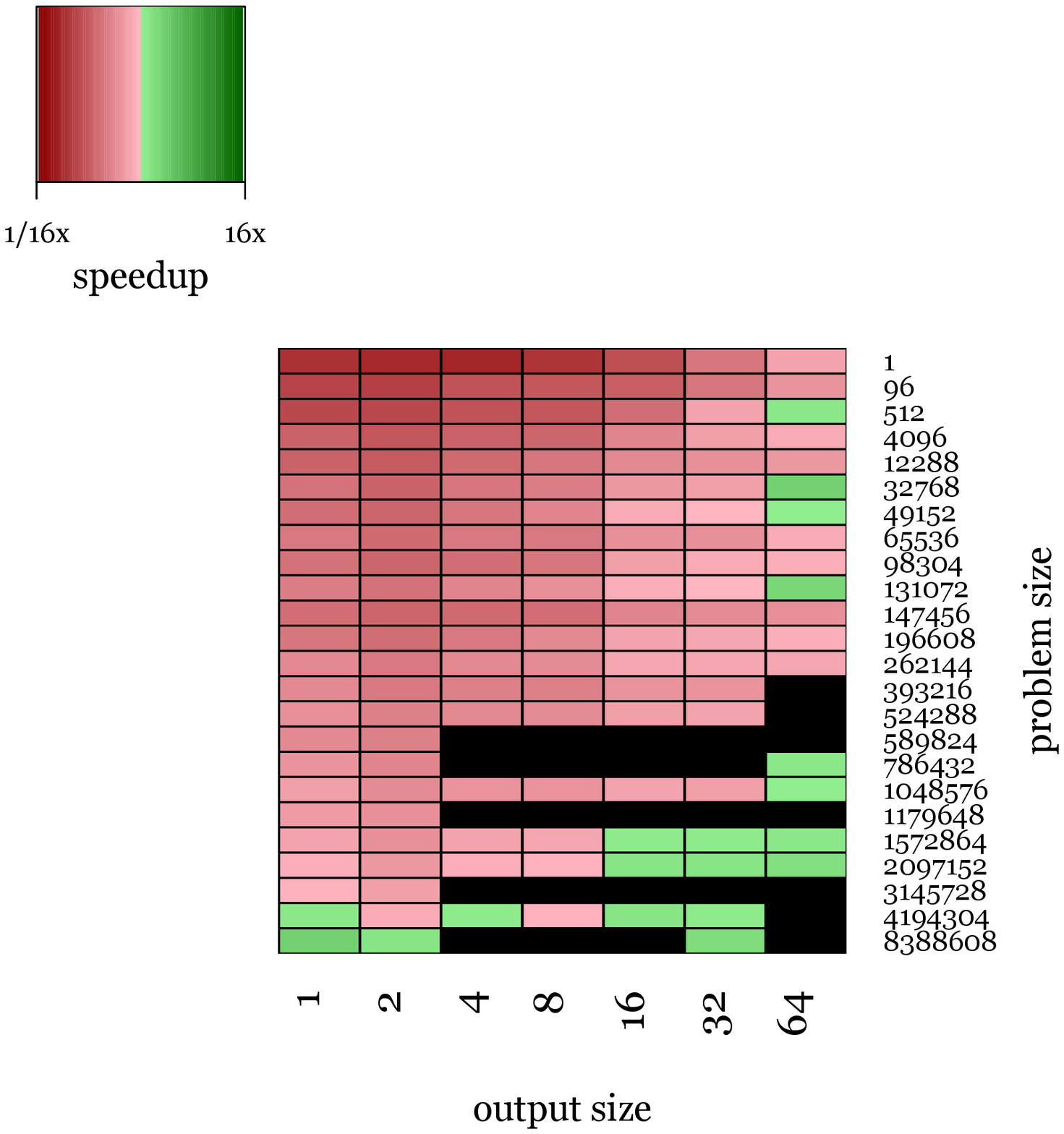}
  \caption{$3 \times 3$ kernel (K40m)}
  \label{fig:ker3}
\end{minipage}%
\begin{minipage}{.5\textwidth}
  \hspace{-1.5cm}
  \includegraphics[width=10.5cm]{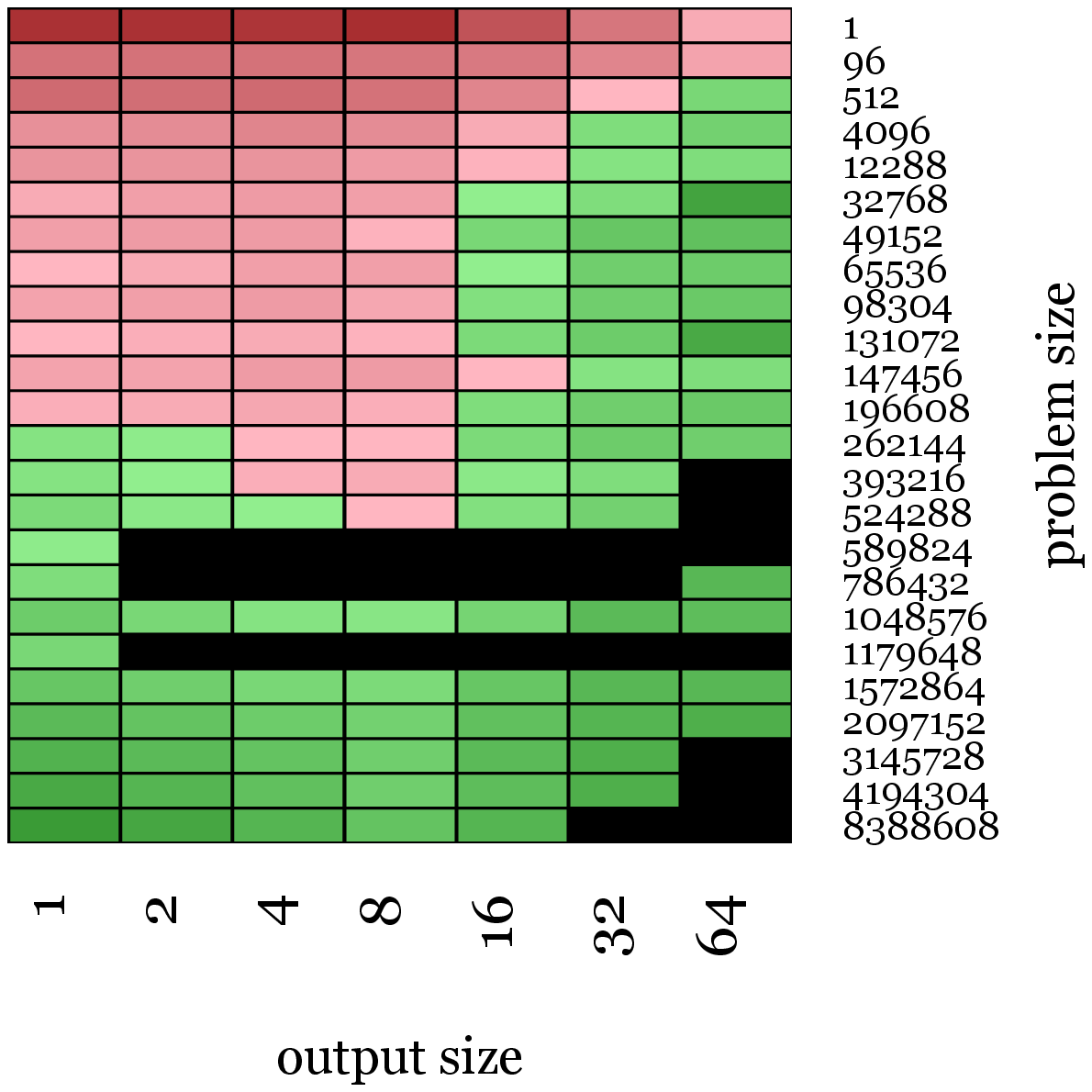}
  \caption{$5 \times 5$ kernel (K40m)}
  \label{fig:ker5}
\end{minipage}%

\vspace{-7.3cm}

\begin{minipage}{.5\textwidth}
  \hspace{-1cm}
  \includegraphics[width=10.5cm]{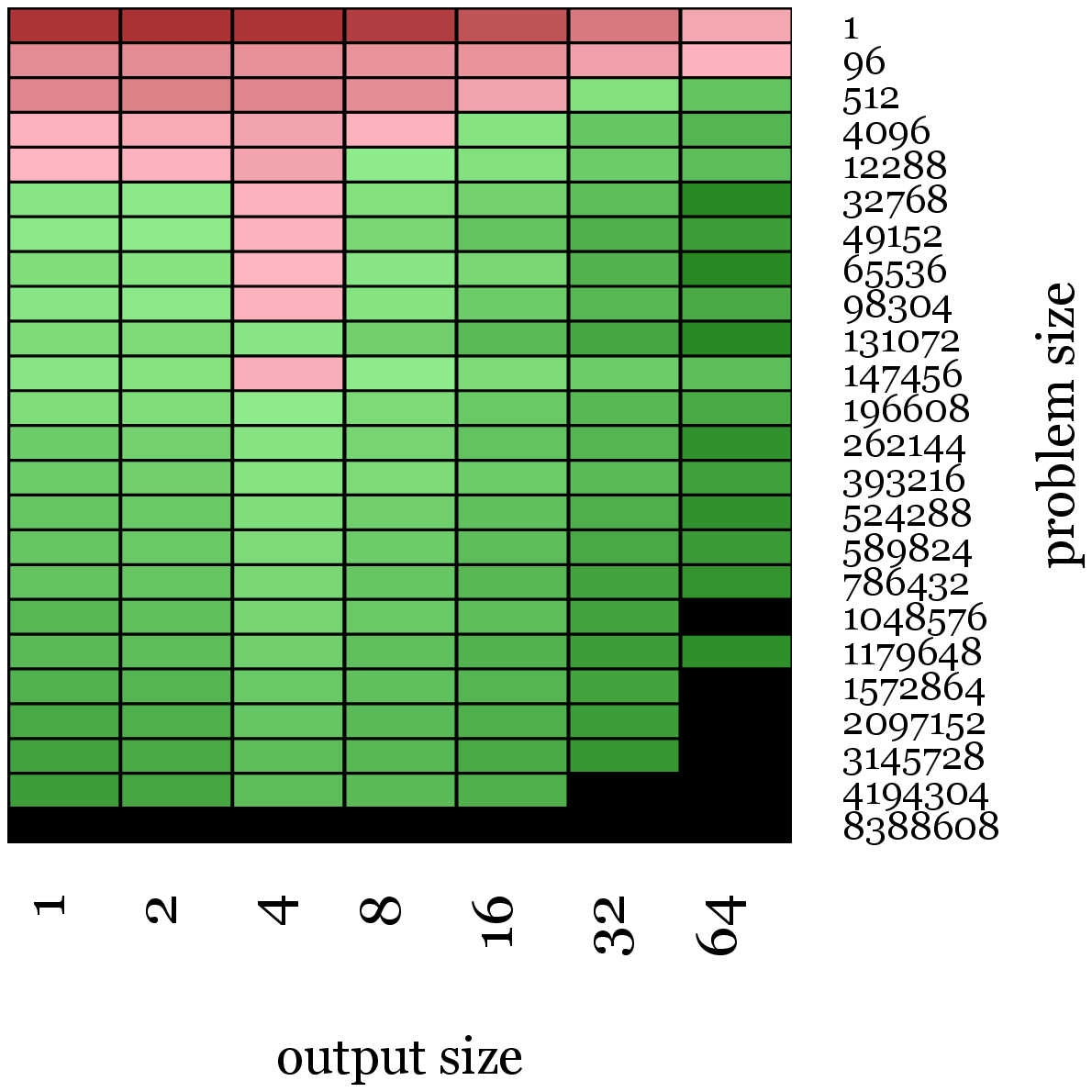}
  \caption{$7 \times 7$ kernel (K40m)}
  \label{fig:ker7}
\end{minipage}%
\begin{minipage}{.5\textwidth}
  \hspace{-1.5cm}
  \includegraphics[width=10.5cm]{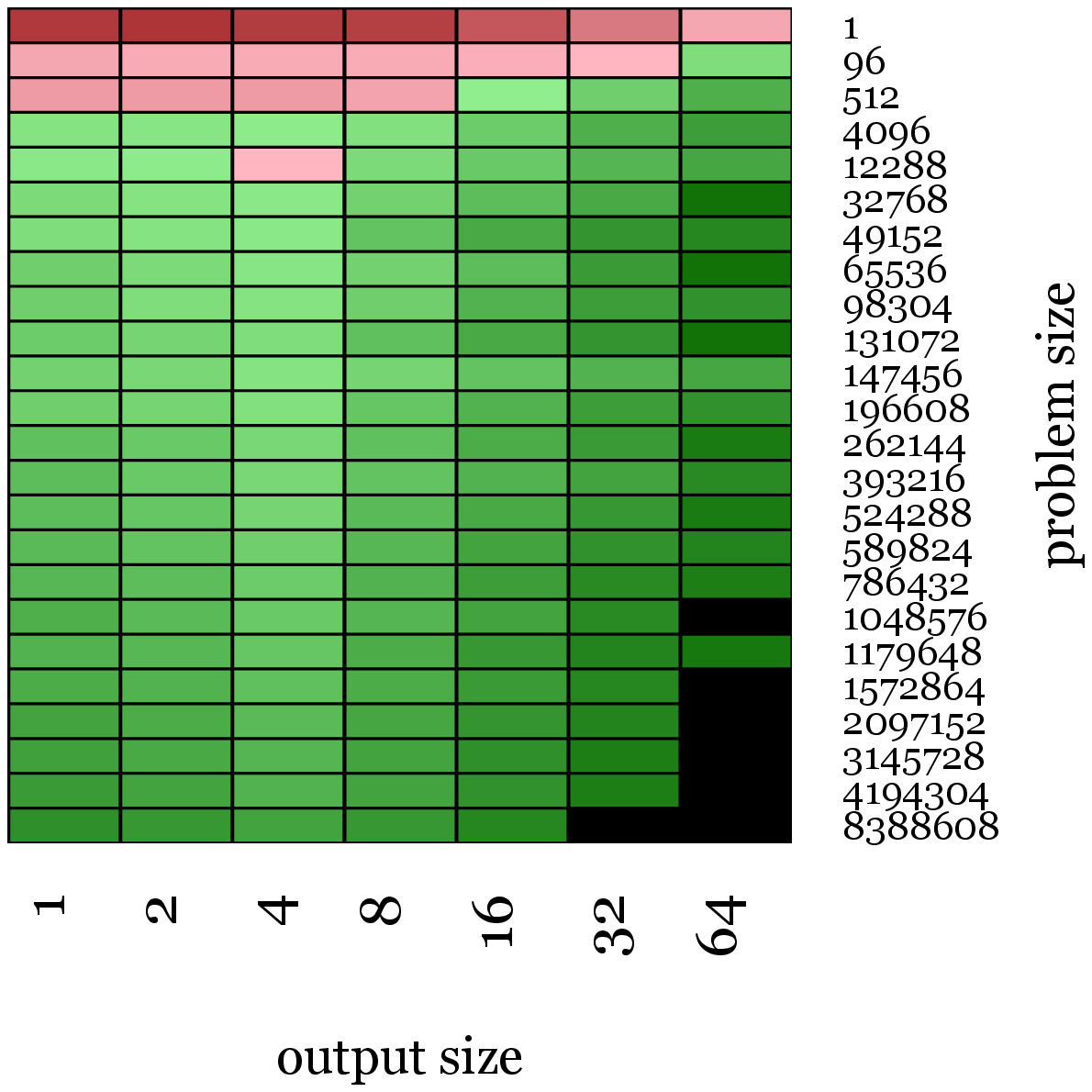}
  \caption{$9 \times 9$ kernel (K40m)}
  \label{fig:ker9}
\end{minipage}%

\vspace{-7.3cm}

\begin{minipage}{.5\textwidth}
  \hspace{-1cm}
  \includegraphics[width=10.5cm]{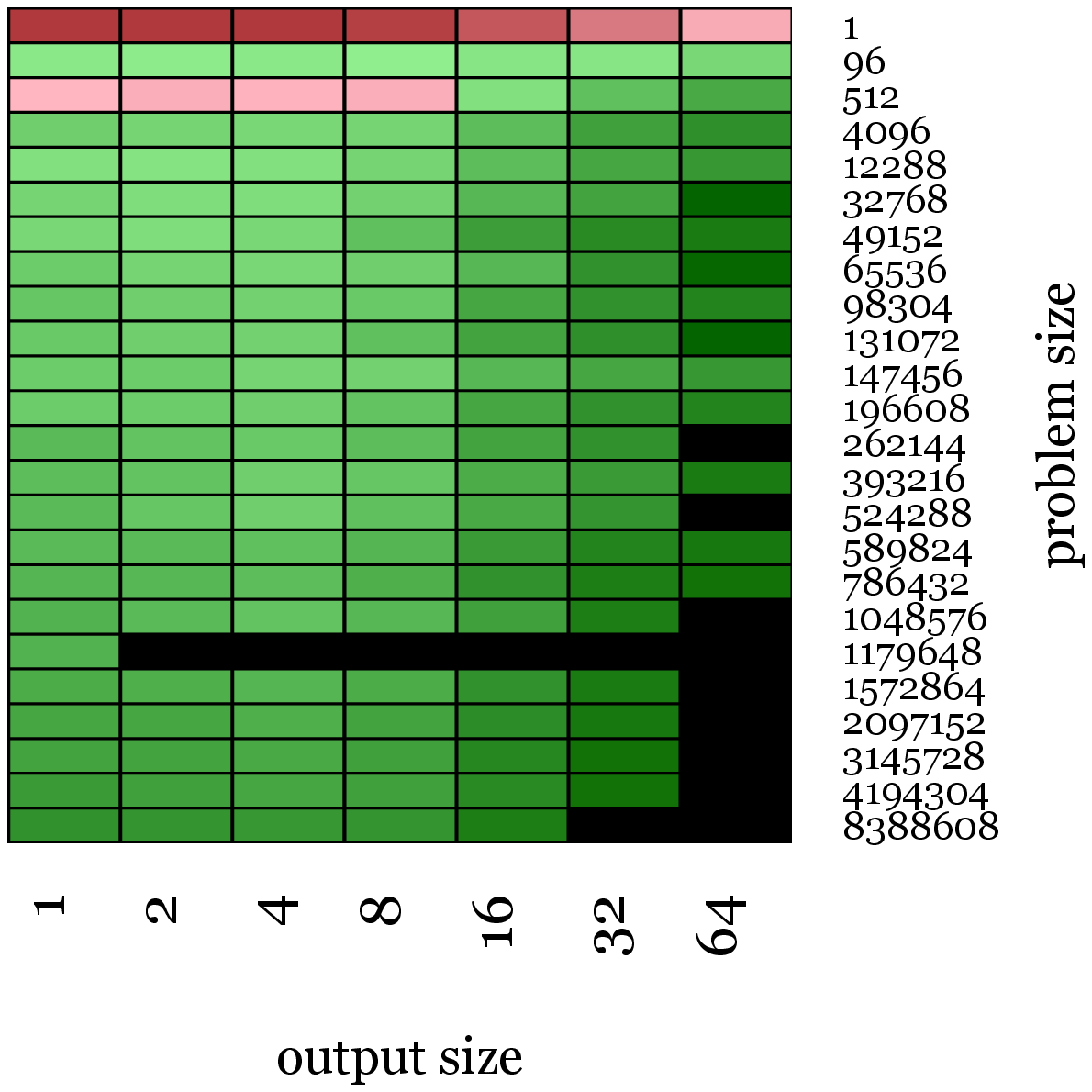}
  \caption{$11 \times 11$ kernel (K40m)}
  \label{fig:ker11}
\end{minipage}%
\begin{minipage}{.5\textwidth}
  \hspace{-1.5cm}
  \includegraphics[width=10.5cm]{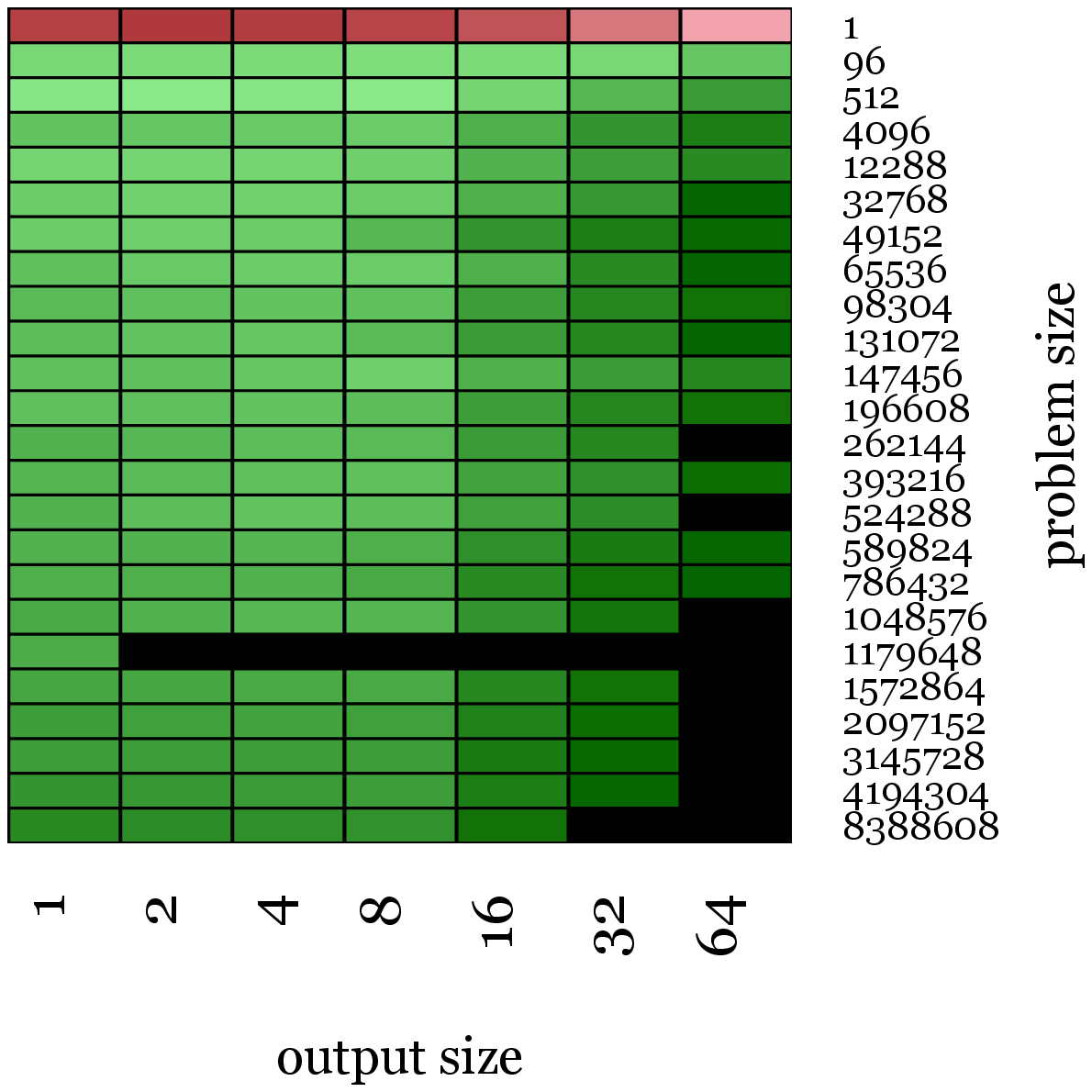}
  \caption{$13 \times 13$ kernel (K40m)}
  \label{fig:ker13}
\end{minipage}%
\end{figure}

Figures \ref{fig:ker3}-\ref{fig:ker13} are performance summaries of \cufft convolution
versus \cudnn on a \nvidia Tesla K40m, averaged across all three passes. The $y$-axis \emph{problem size} corresponds
to the minibatch size multiplied by number of input and output planes ($S f f'$); each one of these
is a pass reduction dimension. Many possible combinations of $S, f, f'$ may map to the same problem size.
\cudnn performance varies to a greater degree than \cufft across passes.
This is due to the asymmetry of convolution
sizes in each pass, and the fact that a larger convolution kernel (as seen with gradient accumulation)
is essentially free in the Fourier domain. Averaging the three passes together provides a proxy for
overall performance.
The $x$-axis corresponds to output height/width.
For deeper layers in image \cnnsend, output size will decrease while $f, f'$ will increase,
so depth corresponds to moving  from the upper right to the lower left of the graph.
Black areas in the chart are due to failed \cufft runs, due to memory pressure
or undetermined potential \cufft 6.5 issues.

\fft convolutions make large kernel sizes inexpensive, which make the performance of all three
passes roughly equal (Table \ref{tab:fft_perf}).
On the other hand, zero-padding $k_h \times k_w$ to $h \times w$ penalizes smaller
kernels compared to \cudnnend.
For $3\times 3$ kernels (Figure~\ref{fig:ker3}), \cufft performance is poor compared to \cudnnend.
The overhead of multiple kernel launches, streaming memory in and out multiple times,
and zero-padding to the input size often outweigh the algorithmic advantage
of FFT. However, for the largest problem sizes, $3 \times 3$ convolution via
FFT can still be advantageous, with top speed $1.84 \times$ faster than \cudnnend.
$5 \times 5$ kernels (Figure~\ref{fig:ker5}) show an increasing dominance of the FFT strategy, with top
speed $5.33 \times$ faster. The tendency is confirmed for larger kernel sizes:
at $13 \times 13$, maximum speedup is $23.54 \times$ over \cudnnend.

\hspace{-3cm}
\newpage

\subsection{\cnn Performance}
\label{sec:raw_performance}

In table \ref{tab:alexnet_overfeat}, we show performance for real \cnnsend,
AlexNet (\cite{NIPS2012_4824}) and OverFeat \emph{fast}
(\cite{sermanet-iclr-14}), comparing against \cudnn and cuda-convnet2 (ccn2)
kernels in Torch. The first layer uses \cudnn for the \cufft runs because
it is strided, but all other layers use \cufftend. The timings include all
convolutional layers of the network.

\begin{table}[h]
\caption{AlexNet and OverFeat \emph{fast} performance (K40, $ms$)}
\label{tab:alexnet_overfeat}
\begin{center}
\begin{tabular}{llllll}
\multicolumn{1}{c}{\bf NETWORK}
&\multicolumn{1}{c}{\bf KERNEL}
&\multicolumn{1}{c}{\bf FPROP}
&\multicolumn{1}{c}{\bf BPROP}
&\multicolumn{1}{c}{\bf ACCGRAD}
&\multicolumn{1}{c}{\bf TOTAL}
\\ \hline \\
AlexNet & \cufft & $\mathbf{94.34}$ & $\mathbf{96.69}$ & $\mathbf{93.20}$ & $\mathbf{284.23}$ \\
        & \cudnn & $147.32$ & $167.79$ & $153.96$ & $469.07$ \\
        & ccn2   & $99.03$ & $104.59$ & $103.29$ & $306.91$ \\
\hdashline \\
OverFeat \emph{fast} & \cufft & $\mathbf{375.65}$ & $460.48$ & $\mathbf{397.85}$ & $\mathbf{1233.98}$ \\
         & \cudnn  & $459.06$ & $634.26$ & $508.02$ & $1601.35$ \\
         & ccn2    & $433.11$ & $\mathbf{398.87}$ & $450.82$ & $1282.80$ \\
\end{tabular}
\end{center}
\end{table}

Table~\ref{tab:fft_perf} shows the performance of the \cudnn and our
\cufft convolution implementation for some representative layer sizes,
assuming all the data is present on the
\gpuend. Our speedups range from $1.4\times$ to $14.5\times$ over
\cudnnend. Unsurprisingly, larger $h, w$, smaller $S, f, f', k_h, k_w$
all contribute to reduced efficiency with the \fftend.
More surprisingly, we experience noticeable speedups on small $3 \times 3$ kernels
as long as the input tensor remains of small size.
The optimal FFT sizes that autotuning finds are reported in columns $2$ and $3$; note L5
padding being found by the autotuner.
Column $7$ has the trillion \emph{equivalent
time-domain reductions per second} (single-precision floating point multiply-adds)
achieved by our implementation on a \nvidia Tesla K40m on \cuda 6.5.
This number represents the throughput a time-domain kernel needs to
achieve in order to match our implementation; it is computed as
$(S f f' k_h k_w (h-k_h+1) (w-k_w+1)) / time$.
This is a metric to compare relative efficiency across problem and
padding sizes. In the cases L2, L3 and L4, a time domain
convolution would need to exceed the K40m peak of $4.29$ Tflop/sec in order to
match our throughput.

\begin{table}[h]
\caption{Representative layer performance ($S = 128$, K40m)}
\label{tab:fft_perf}
\begin{center}
\begin{tabular}{lllllll}
\multicolumn{1}{c}{\bf LAYER}
&\multicolumn{1}{c}{\bf $h+p_h$}
&\multicolumn{1}{c}{\bf $w+p_w$} &\multicolumn{1}{c}{\bf \cudnn}
&\multicolumn{1}{c}{\bf \cufft} &\multicolumn{1}{c}{\bf SPEEDUP}
&\multicolumn{1}{c}{\bf TRED/$s$}
\\ \hline \\
\multicolumn{1}{l}{\bf L1} &
\multicolumn{6}{l}{Params: $f = 3, f' = 96, h = w = 128, k_h = k_w = 11$} \\
fprop & $128$ & $128$ & $125.11~ms$ & $80.98~ms$ & $1.54\times$ &  $0.9$ \\
bprop & $128$ & $128$ & $153.39~ms$ & $66.49~ms$ & $2.30\times$ &  $1.1$ \\
accGrad & $128$ & $128$ & $155.07~ms$ & $69.63~ms$ & $2.22\times$ & $1.05$ \\
\hdashline
\multicolumn{1}{l}{\bf L2} &
\multicolumn{6}{l}{Params: $f = 64, f' = 64, h = w = 64, k_h = k_w = 9$} \\
fprop & $64$ & $64$ & $354.83~ms$ & $46.44~ms$ & $7.64\times$ & $7.49$ \\
bprop & $64$ & $64$ & $579.37~ms$ & $46.25~ms$ & $12.5\times$ & $7.52$ \\
accGrad & $64$ & $64$ & $416.34~ms$ & $47.03~ms$ & $8.85\times$ & $7.40$ \\
\hdashline
\multicolumn{1}{l}{\bf L3} &
\multicolumn{6}{l}{Params: $f = 128, f' = 128, h = w = 32, k_h = k_w = 9$} \\
fprop & $32$ & $32$ & $130.89~ms$ & $17.77~ms$ & $7.36\times$ &  $9.90$ \\
bprop & $32$ & $32$ & $245.57~ms$ & $16.97~ms$ & $14.5\times$ & $10.37$ \\
accGrad & $32$ & $32$ & $154.96~ms$ & $17.00~ms$ & $9.29\times$ & $10.34$ \\
\hdashline
\multicolumn{1}{l}{\bf L4} &
\multicolumn{6}{l}{Params: $f = 128, f' = 128, h = w = 16, k_h = k_w = 7$} \\
fprop & $16$ & $16$ & $15.13~ms$ & $4.88~ms$ & $3.10\times$ & $5.54$ \\
bprop & $16$ & $16$ & $20.80~ms$ & $4.71~ms$ & $4.41\times$ & $5.76$ \\
accGrad & $16$ & $16$ & $18.17~ms$ & $4.70~ms$ & $3.86\times$ & $5.75$ \\
\hdashline
\multicolumn{1}{l}{\bf L5} &
\multicolumn{6}{l}{Params: $f = 384, f' = 384, h = w = 13, k_h = k_w = 3$} \\
fprop & $13$ & $14$ & $39.82~ms$ & $21.35~ms$ & $1.86\times$ & $1.34$ \\
bprop & $13$ & $14$ & $28.33~ms$ & $20.22~ms$ & $1.40\times$ & $1.42$ \\
accGrad & $13$ & $14$ & $47.84~ms$ & $21.26~ms$ & $2.25\times$ & $1.35$ \\
\end{tabular}
\end{center}
\end{table}

\section{\fbfft Implementation}
\label{sec:fbfft}

This section presumes familiarity with \gpu architecture. Refer to the Supplement for details.

When designing high-performance libraries, multiple objectives must be
balanced against each other: memory latency/bandwidth tradeoffs,
maximizing locality without sacrificing too much parallelism, good instruction
mix, register usage and mapping strategy of computation and data to memories
and compute elements. A key principle is to design a set
of leaf kernels with well-tuned in-register performance and reduce the larger
problem to a combination of these kernels by data and loop
tiling (\cite{Irigoin}) and recursive decompositions (\cite{Gunnels2001}).
Since vendors have to sustain high performance for a large class of application
domains, there exist parameter configurations for which a
carefully tuned approach significantly outperforms vendor-tuned
libraries (\cite{Shin:2010}).
For common deep learning use, convolutional layers consist of many
batched small 2-D convolutions. These are tiny relative to DSP
and HPC standards and put us in a regime where (a) we fall outside of the
highly tuned regime, (b) feature dimensions are often smaller than GPU warp sizes
and can often fit exclusively in registers rather than in shared memory (SMEM),
and (c) we are very sensitive to latencies. We determined that it is possible to obtain
better efficiency than the existing batched \cufft mode for \cnnsend.

\subsection{Limitations of \cufft}
Because the \cufft library is a black box, zero-padding\footnote{This is
different from the FFTW compatibility padding mode for in-place transforms.} has
to be explicitly embedded in the input and output arrays.
The consequence is that one may need to allocate a duplicate, larger memory
region (only once) and copy data from non-padded tensors to padded tensors.
This memory consumption and spurious copies affect latency significantly.
Instead, we devised an implementation for batched \onedfft and \twodfft of sizes
2-256 and \emph{reaches up to 78\% efficiency at 97.5\% occupancy}.
We also implemented an \ifft kernel based on our \fft kernel.

In our implementation we use \emph{clipping} to conditionally load a value if
reading within bounds or a constant ($0$) otherwise.
This is an approach used in automatic code generation tools such as
Halide (\cite{Halide}) and relies on aggressive if-conversion properties of the
\cuda compiler.
It allows for more efficient control flow rather than using explicit loop
prologues and epilogues.
This mechanism does not require any additional memory
allocation and is zero-copy; this is particularly desirable in the latency
sensitive mode.

Additionally, since \cufft and \cublas are closed source, it is impossible to take
advantage of algorithmic simplifications that may be available.
For instance, in the forward pass of our computation as shown in
Table~\ref{tab:passes}, the result of the first \cufft call is of the form
$S\times f \times (h+p_h) \times (\floor{(w+p_w)/2}+1)$.
With \fbfft we return it in the form
$S\times f \times (\floor{(w+p_w)/2}+1) \times (h+p_h)$
where the two innermost data dimensions are transposed. This allows
us to remove a full data transposition from each of the \fft kernels.
Another domain-specific optimization we have yet to explore is eliminating
bit reversal portions of the \fft and \ifftend. This can be done
by performing the \fft with \emph{decimation in frequency} (DIF) and the \ifft with
\emph{decimation in time} (DIT), discussed in the Supplement.

\subsection{Warp-level \onedfft and \twodfft for size $n \leq 32$}
For batched \fft of power of two sizes we view a single warp as a small
distributed system with lockstep collective communication capabilities and we program it
in a bulk-synchronous fashion (\cite{BSP}).
We implement \dif and enforce the following invariants for the $\log_2 n$ steps:
\begin{itemize}
\item each warp thread originally loads one real element of the input vector
and locally computes one complex twiddle factor (i.e. a root of unity);
\item at each step, all warp threads exchange data with
another thread in the warp in parallel and produce a new value;
\item then, all warp threads exchange twiddle factors with another
  thread in the warp in parallel, and produce a new value.
\end{itemize}
The two bulk-synchronous exchanges can be written each with one warp-wide
instruction.
After the $\log_2 n$ steps, the \fft is computed and stored in a
distributed and bit reversed manner within $1$ register across a warp.
For sizes $n \leq 32$, bit reversal can be implemented with a single warp shuffle.

We either load twiddle factors from device memory or compute them with the
{\tt sincosf} function \emph{only once}, and subsequently swap them within registers.
This greatly reduces the reliance on either memory bandwidth or on the special
functional unit at the expense of a few additional registers.
The decision between explicitly loading twiddle factors from
device memory or computing them is a tradeoff between arithmetic
intensity and memory bandwidth.
For sizes $16$ and $32$ the arithmetic pipeline is the
bottleneck. Loading twiddle factors from memory for these two special sizes
results in a performance increase of $15\%$ and $20\%$ respectively.

The discussion above applies to \onedfft and to each independent \fft within a
larger \twodfftend.
A $n$-D Fourier transform is separable and can be implemented with
sets of multiple \onedfft with transpositions between each of these sets.
In \twodfft $\mathbb{R}$-to-$\mathbb{C}$, the first set comprises $n$ \ffts and
the second set comprises $n/2+1$ \ffts by Hermitian symmetry.
Following standard techniques~\cite{Lyons1996} we further pack $2$ real \ffts
into a single complex \fft.
The extra $1$ term in the quantity $n/2+1$ makes the computation ill-balanced
and can bring down performance by lowering occupancy.
We chose to dimension our kernels to have size $n\times(n/2)$ and introduce
additional control flow to handle the border case. This results in $30\%$
additional performance.
We implement the transposition in SMEM across warps following
\cite{PauliusTranspose}. Data is
already resident in registers so our main concerns are limiting SMEM usage to
keep occupancy high, and limiting load/stores by using vector instructions
to avoid saturating the load-store unit (LSU).

\subsection{\onedfft and \twodfft for size $32 < n \leq 256$}
With size $32$ as our building block, we extend our strategy to
larger sizes.
We use the same single warp approach to compute a full \onedfftend.
The main difference is that the computation is now distributed across multiple
registers across threads in a warp ($\ceil{n / 32}$ Fourier coefficients and
twiddle factors in registers per thread).
Because we perform a full \fft per warp, a performance
cross-over where \cufft wins happens after register usage limits occupancy too much.
We outperform 1-D \cufft for $n \leq 256$, with a hard register limit at $n=512$
($128$ and $256$ similarly for \twodfftend). This is still well within our application domain.
The following modifications handle multiple registers per thread:
\begin{itemize}
\item Hermitian symmetry allows us to perform half the computation. There
  is a tradeoff between adding control-flow divergence and
  performing less work. At $n \geq 64$, benefits from reduced
  computations dominate divergence losses;
\item we take advantage of trigonometric symmetries and twiddle
  factor distribution to compute only a fraction of the roots of unity needed
  for each \fftend, distributed with register to register copies;
\item twiddle factor re-balancing across a warp and across registers requires a
  different implementation. We managed to implement it fully within registers;
\item bit reversal occurs across registers and across warps. The high-order
  bits represent the register while the low-order bits represent the warp.
  Without a sophisticated implementation,
  this results in indirect addressing of registers which is costly.
  We implement a simple bit reversal in SMEM,
  which is an occupancy bottleneck at $n \geq 256$ for \onedfftend.
\end{itemize}
In the \twodfft case, the intermediate transpose becomes significantly
more expensive. We experimented with various strategies to keep occupancy high, including partial
transpositions within a warp to use minimal amounts of SMEM.

\subsection{Discussion}

We report the relative performance of our implementation \fbfft compared to
\cufft for various batch and input sizes of interest.
The number of batches to consider depends on the dimension of \cnn layers
as well as any multi-\gpu parallelization strategy that may be involved.
At typical sizes of interest, \fbfft is between $1.5\times$ and $5\times$ faster.
We tried up to $4$ million batches and at larger
sizes gains stabilize around $1.4\times$ but efficiency goes down
as more and more memory is used.

\begin{figure}[h]
\begin{center}
\begin{minipage}{.5\textwidth}
  \includegraphics[width=\textwidth]{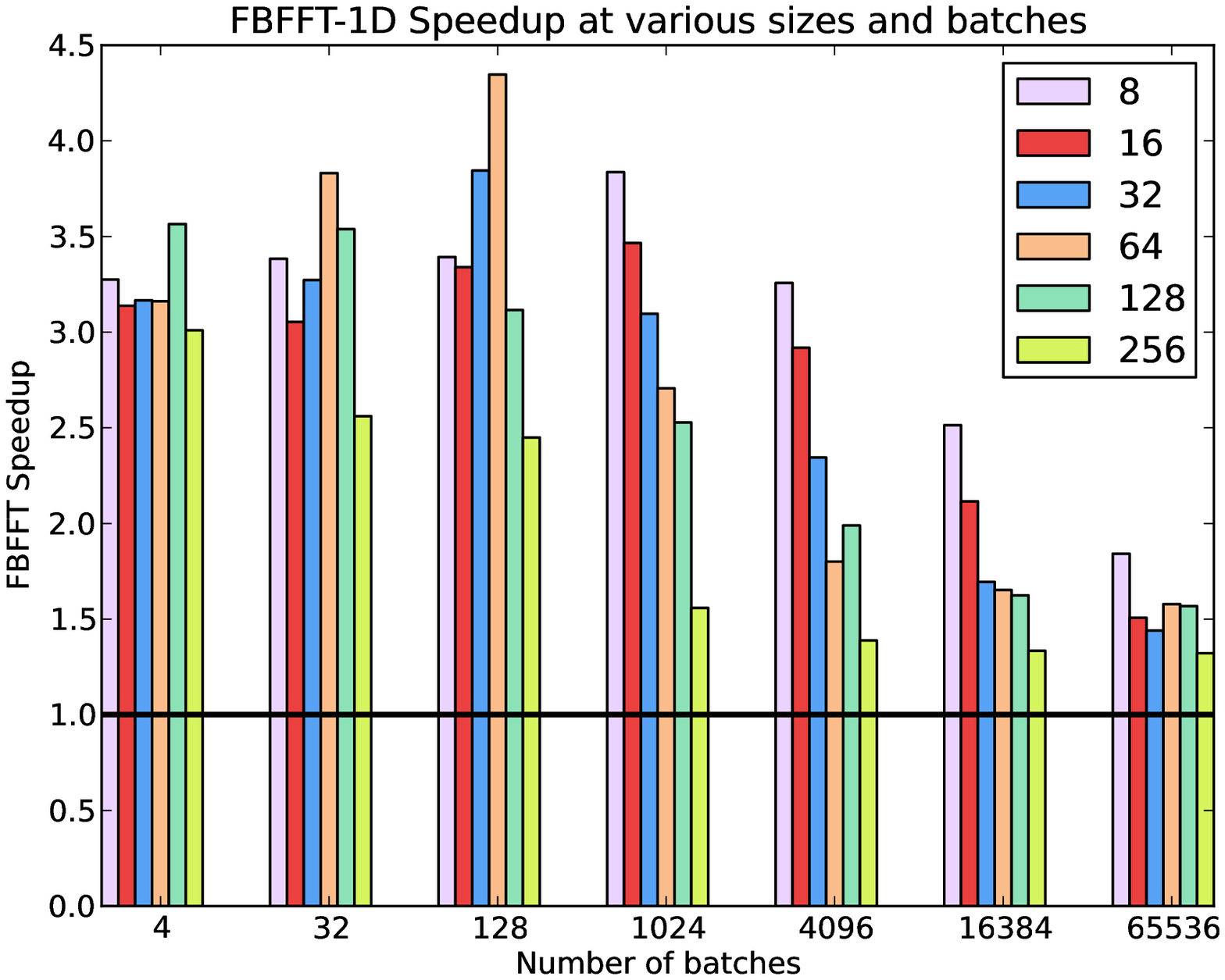}
\end{minipage}%
\begin{minipage}{.5\textwidth}
  \includegraphics[width=\textwidth]{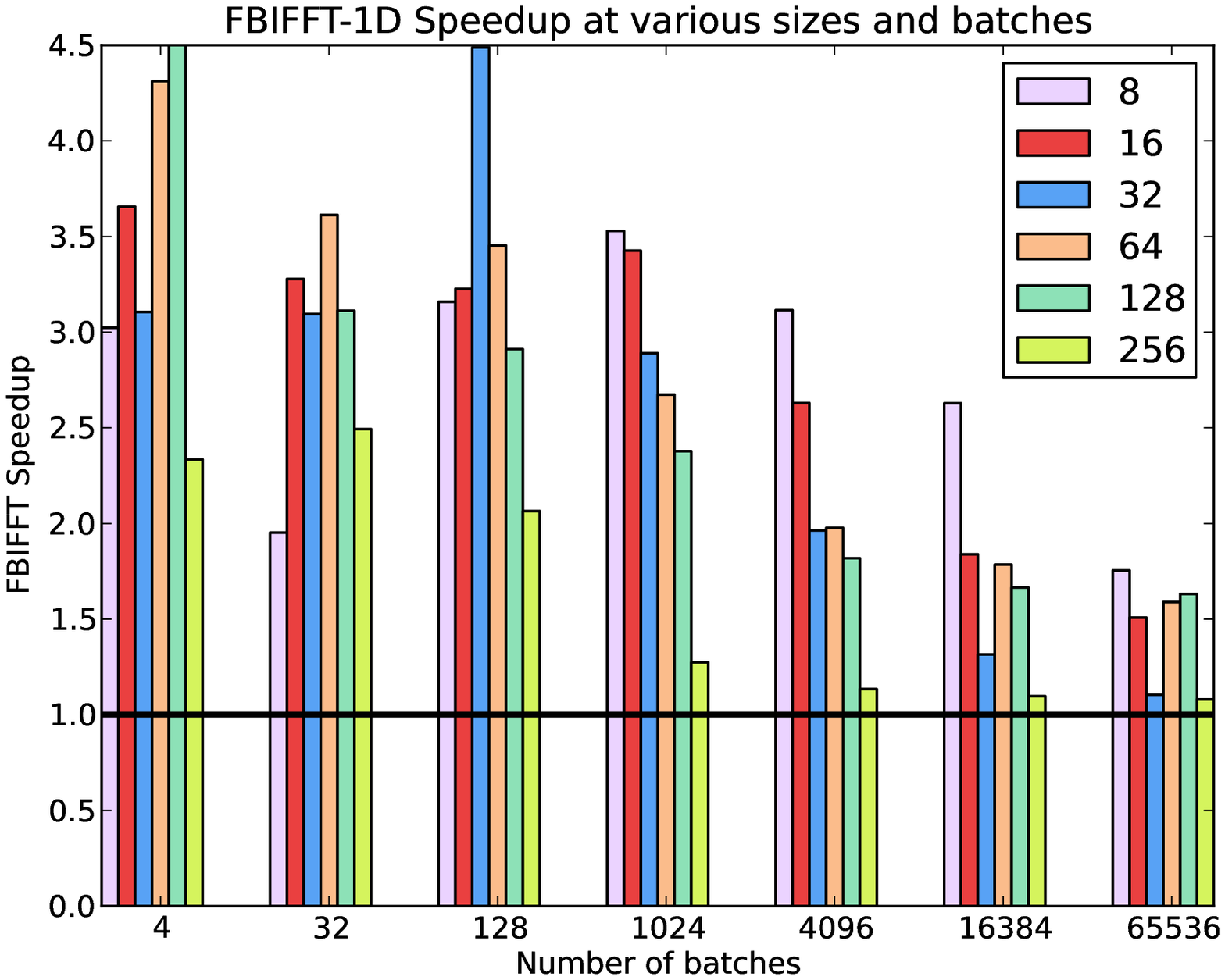}
\end{minipage}
\caption{\fbfftend-1D \fft and \ifft (K40m, \cufft 6.5 @ 1x)}
\label{fig:fft1d}

\begin{minipage}{.5\textwidth}
  \includegraphics[width=\textwidth]{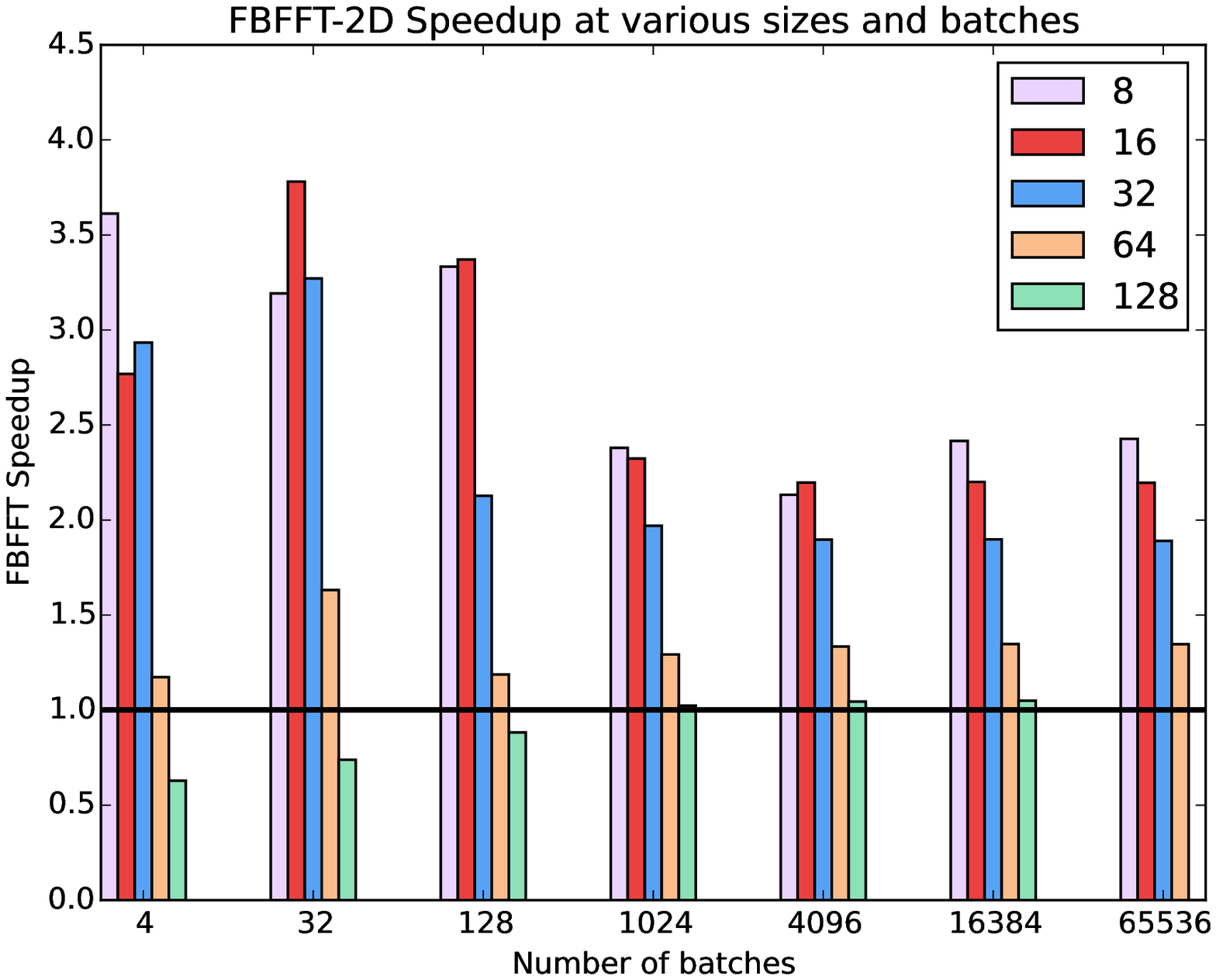}
\end{minipage}%
\begin{minipage}{.5\textwidth}
  \includegraphics[width=\textwidth]{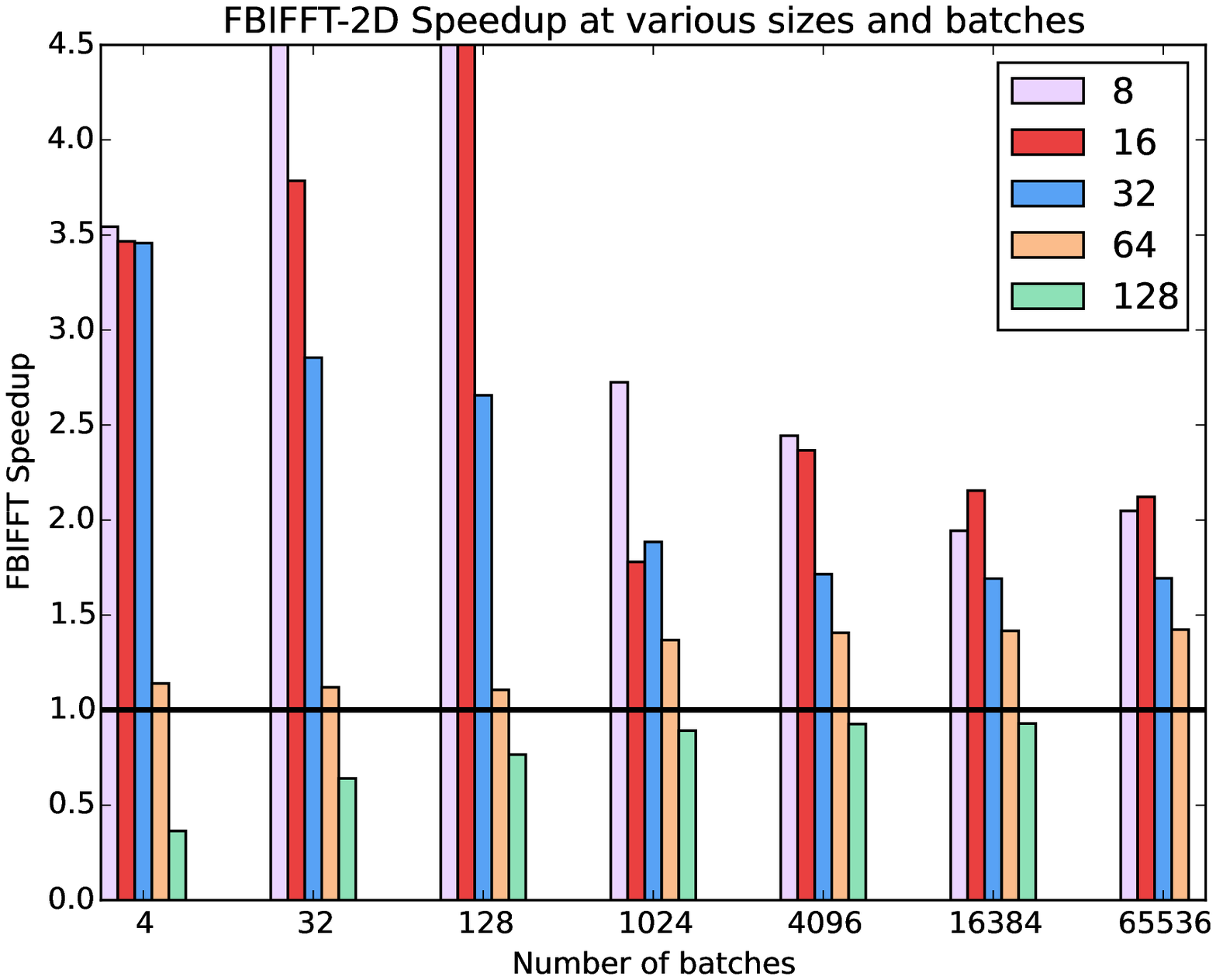}
\end{minipage}
\caption{\fbfftend-2D \fft and \ifft (K40m, \cufft 6.5 @ 1x)}
\label{fig:fft2d}
\end{center}
\end{figure}

Figure~\ref{fig:fft1d} shows the performance in the 1-D case.
These numbers do not exercise our implicit zero-copy padding, so we expect
additional gains when we incorporate our \fft in the convolution.
Our implementation outperforms \cufft for all cases of interest, more
dramatically so for smaller batch sizes.
Small batch sizes also correspond to the latency sensitive regime in
Figures~\ref{fig:ker3}-\ref{fig:ker13} for which the \cufft based
implementation performs quite worse than \cudnnend.
\emph{We achieve $78\%$ efficiency at $97.5\%$ occupancy for size $64$
at batch size $16,384$}, as reported by {\tt nvvp}.

Figure~\ref{fig:fft2d} shows the performance in the 2-D case.
Relative performance gains for sizes $64$ are more modest than in the 1-D
case, even losing to \cufft at size $128$ and small batch sizes.
The magnitude of the relative gains at various batch sizes drops
faster than in the 1-D case. Looking at the performance of the
$32\times 32$ \fftend, we obtain $1.6 \times$ speedup over \cufft at $1,024$
batches. The same ratio is not obtained until $16,384$ batches in
\onedfftend.\footnote{This is not unexpected because these two computations
  perform the same number of flops when accounting for Hermitian symmetry, plus
  the fact that the efficiency of \cufft increases while \fbfft remains high but
  almost constant.}
When coupled with the tiling strategy in Section \ref{sec:future}, we
emphasize that the sizes of interest are actually $8$-$64$,
and depend on $k_h, k_w$ but not input $h, w$. Batch sizes can vary on the
whole spectrum.

We interfaced \fbfft into our convolution module and ran
experiments with $3\times 3$ kernels for the $3$ different convolution passes
over inputs of
sizes $x = h = w, x \in \{13, 16, 27, 32, 57, 64\}$. For problem size, we used
$p = S = f = f', p \in \{16, 32, 64, 128\}$.
\emph{By swapping our \fft implementation we observed an overall mean speedup
of $1.51\times$ with standard deviation $0.21$ and geometric mean
$1.49\times$.}
The minimum speedup was $1.21\times$, despite sometimes performing more
computations with \fbfft which can only interpolate to a power of $2$.
These experiments exercise the zero-copy padding and lower memory footprints of
\fbfft compared to \cufft but do not yet reflect additional optimizations such
as tiling and bit twiddling elision.

\section{Current Limitations and Future Work}
\label{sec:future}

In our current implementation, \fbfft heavily relies on shuffle
instructions. In spite of a good efficiency, we only utilize $60\%$ of the
available memory bandwidth. This is due to the load and store instructions in
our kernel competing with the shuffle instructions for the Load-Store Unit
(LSU). As a consequence, our first bottleneck is the number of instructions
issued on the LSU. For instance, on Kepler (capability $3.5$), the throughput
for $32$-bit floating point multiply-add operations is $192$ per cycle but the
throughput for shuffles is only $32$. In the future we will investigate and
release faster implementations as they become available.

Temporary memory overhead requirements are a common issue when performing
convolutions in the Fourier domain. In this first implementation, we
introduced the following memory buffers to support our implementation:
\begin{itemize}
\item for each of input, output and weight tensors we store $1$ buffer for the
frequency array and $1$ buffer for its complex transpose. These buffers store
the Fourier representation and are generally limited by the weight tensor
which is independent of the mini-batch size. Because of the global memory pressure
we introduce, we reuse buffers at each layer and pass on the opportunity to
(1) reuse 2 FFT results in each hidden layer, reducing the cost of forward
FFTs by $33\%$; and (2) asynchronously precompute FFTs of the weight tensors
and their gradients to better fill the gpu utilization pipeline,
\item when using \cufft we additionally pad the input, weight and output
  tensors explicitly to the best performing common fft size
\item when using \cufft additional temporary memory is reserved by each
  cufftPlan
\item with \fbfft padding is implicit but and no temporary memory buffer is
  needed until we reach size $64$. On the other hand, \fbfft only supports
  square convolutions whose size is a power of $2$. As a consequence, too much
  padding could occur and adversely affect both performance and memory
  consumption. The tiling strategy we describe next is a good way to
  circumvent the problem.
\end{itemize}
Additionally, we recently developed an in-place transposed batched CGEMM which
permits the removal of the complex transposed buffer. For this problem, a tool
like MaxAS~\cite{DBLP:journals/corr/Lavin15} could be valuable.

\fbfft provides the most gains over \cufft at sizes $8$-$64$. A tiling strategy
for the input can be used to exploit this advantage. When the kernel is
significantly smaller than the input, we can decompose a large convolution into
several smaller ones. For simplicity, we consider 1D convolution
on a single input plane, as it can trivially be extended. Let $x$ be an
input of size $n$, $c$ a kernel of size $w$ and $y=x\star c$. We write $x_{[i,j]}$ for the
vector formed by contiguous elements of $x$: $\{x_i, x_{i+1}, ..., x_{j-1}\}$.
Let $d \le n$. From the definition of the convolution, we have:
$$y_{[i,i+d]} = x_{[i,i+d+w]} \star c$$
So the convolution of the input of size $n$ can be computed with $\lfloor n/d \rfloor$
convolutions with inputs of size $d+w$.
The cost of the convolution goes down from $\mathcal{O}(n\log(n))$ to
$\mathcal{O}(\lfloor n/d \rfloor (d+w)\log(d+w)) = \mathcal{O}((n+w/d)\log(d+w))$.
From this formula, we see that the optimal $d$ is of the order of $w$,
to get the complexity $\mathcal{O}(n\log(w))$. This strategy allows us to speed up
forward and backward propagation.
Tiling can also be used to reduce memory cost for temporary storage by not running all
the tiles in parallel (just the tiles which do run in parallel need their scratch space),
at the potential expense of parallelism or efficiency.

For the gradient accumulation, we cannot reuse this strategy, since it involves a larger
convolution between an input $x$ of size $n$ and a kernel $z = \frac{\partial L}{\partial y}$ of size $n-w+1$.
However, we have a similar formula:

$$\left(\frac{\partial L}{\partial c}\right)_j = \sum_{i=0}^{n-1} x_{j+i}\cdot z_i =
\sum_{k=0}^{\left\lfloor n/d\right\rfloor-1} \sum_{i=0}^{d-1} x_{j+i+kd} \cdot z_{i+kd}
+\sum_{i=d\left\lfloor n/d\right\rfloor}^{n-1} x_{j+i} \cdot z_i$$
And so
$$\left(\frac{\partial L}{\partial c}\right) = \sum_{k=0}^{\left\lfloor n/d\right\rfloor-1} x_{[dk, (d+1)k+w-1]}
\star z_{[dk,(d+1)k]}
+ x_{[d\left\lfloor n/d\right\rfloor, n]} \star z_{[d\left\lfloor n/d\right\rfloor, n-w+1]}$$

We have a few other optimizations that are planned as well. Since much of the data we have is already
available in registers or in shared memory, we are implementing our own in-place, in-register transpose
via recursive decomposition. The pointwise multiplications in the Fourier domain, especially with tiling,
are rather small, so our own matrix multiplication routines integrated with the rest of the
convolution kernel code might win over \cublasend, and prevent the need for multiple \cuda kernel launches
and their associated overhead.
Finally, as mentioned earlier, bit reversal portions can be eliminated with the FFT using
DIF and the IFFT using DIT.

\section{Conclusion}
To summarize, we achieve significant gains in \cnns using \fftsend, with
a \cufft convolution implementation achieving
$1.4\times - 14.5\times$ speedups over \cudnn for common sizes. In reaction
to \cufft and \cublas limitations in the context of our specific application
domain, we developed our own \fft implementation, \fbfftend, which is more
suited to deep learning problem sizes (large batches, small feature planes).
\fbfft itself is $\geq 1.4\times$ faster than \cufft transforms for these problems of interest.
For convolution, it is faster than the \cufft as well, with a mean of $1.51\times$ for sizes that we wish to
exploit.

Given our new efficient primitive for size $8$-$64$ convolution, we are continuing work on
bit twiddling, transposition and pointwise multiplication optimizations, and continuing
work on tiling to make the computational advantage at that size apply to larger convolution
problems. These will all allow for reduced training time and use of ever larger and
deeper \cnnsend.

\subsubsection*{Acknowledgments}
We would like to thank Julien Demouth from NVIDIA who suggested further
improvements are still possible by virtue of the current implementation being
LSU throughput-bound rather than memory-bound.

\bibliography{paper}
\bibliographystyle{iclr2015}

\newpage
\section{Supplement}
\label{sec:supplement}

\subsection{\cufft Convolution Performance Breakdown}
We show a breakdown of \cufft convolution performance for the steps
indicated in Table~\ref{tab:passes}.
The timings do not add up to $100\%$ of the reported performance in the
previous table because we do not report additional copies needed for
zero-padding here. We also enforce force extra synchronizations to isolate the
contribution of each operation.
Abstracting from these details, the \fft and \ifft take up a significant
amount of compute resources, which we address in Section~\ref{sec:fbfft}.

\begin{table}[h]
\caption{\cufft convolution performance breakdown (K40m, $ms$)}
\label{tab:breakdown}
\begin{center}
\begin{tabular}{llllllll}
\multicolumn{1}{c}{\bf LAYER}  &\multicolumn{1}{c}{\bf FFT A}
&\multicolumn{1}{c}{\bf TRANS. A} &\multicolumn{1}{c}{\bf FFT B}
&\multicolumn{1}{c}{\bf TRANS. B} &\multicolumn{1}{c}{\bf CGEMM}
&\multicolumn{1}{c}{\bf TRANS. C} &\multicolumn{1}{c}{\bf IFFT C}
\\ \hline \\
\multicolumn{8}{l}{\bf L1} \\
fprop & $0.86$ & $0.24$ & $1.13$ & $0.32$ & $15.13$ & $12.67$ & $36.46$ \\
bprop & $0.86$ & $0.24$ & $34.55$ & $10.26$ & $12.62$ & $0.39$ & $1.19$ \\
accGrad & $1.14$ & $0.32$ & $34.60$ & $10.26$ & $12.37$ & $0.26$ & $0.91$ \\
\hdashline
\multicolumn{8}{l}{\bf L2} \\
fprop & $2.99$ & $0.98$ & $5.91$ & $2.03$ & $8.92$ & $1.67$ & $6.24$ \\
bprop & $2.99$ & $0.98$ & $5.92$ & $2.03$ & $8.85$ & $1.67$ & $6.23$ \\
accGrad & $5.94$ & $2.04$ & $5.93$ & $2.02$ & $8.38$ & $0.83$ & $3.15$ \\
\hdashline
\multicolumn{8}{l}{\bf L3} \\
fprop & $3.07$ & $0.89$ & $3.08$ & $0.89$ & $4.40$ & $0.87$ & $3.49$ \\
bprop & $3.08$ & $0.89$ & $3.07$ & $0.90$ & $4.05$ & $0.86$ & $3.48$ \\
accGrad & $3.07$ & $0.89$ & $3.06$ & $0.89$ & $4.03$ & $0.87$ & $3.48$ \\
\hdashline
\multicolumn{8}{l}{\bf L4} \\
fprop & $0.84$ & $0.24$ & $0.83$ & $0.24$ & $1.21$ & $0.24$ & $0.95$ \\
bprop & $0.83$ & $0.24$ & $0.83$ & $0.24$ & $1.13$ & $0.23$ & $0.94$ \\
accGrad & $0.84$ & $0.24$ & $0.82$ & $0.24$ & $1.10$ & $0.24$ & $0.95$ \\
\hdashline
\multicolumn{8}{l}{\bf L5} \\
fprop & $7.07$ & $1.58$ & $2.39$ & $0.51$ & $6.23$ & $0.50$ & $2.54$ \\
bprop & $7.07$ & $1.59$ & $2.40$ & $0.51$ & $5.59$ & $0.51$ & $2.54$ \\
accGrad & $2.40$ & $0.51$ & $2.38$ & $0.52$ & $6.18$ & $1.54$ & $7.51$ \\
\end{tabular}
\end{center}
\end{table}

In the particular case of L1, the FFTs take more than $50\%$ of the runtime.
This is due to the wasteful interpolation of the kernel tensor from a
$11\times 11$ up to $128\times 128$, which is the minimal size to compute the
FFT of the input array without interpolation loss. In such cases, the tiling
strategy we are developing (see section~\ref{sec:future}) will result in large
additional performance gains.

\subsection{\fft: Decimation in Time vs Frequency}
\label{subsec:decimation}
A Fourier transform projects $\mathbb{R}$ and $\mathbb{C}$-valued functions onto a harmonic
orthogonal basis.
The discrete Fourier transform of a vector $\{x_k\},~k\in[0, n-1]$ is the vector:
$$ \{X_k\} = \left(\sum_{j=0}^{n-1} x_j w_{n}^{kj}\right),~k\in[0, n-1]$$
where $w_{n}^j~=~e^{-2\pi i j/n}$ is the $j^{th}$ $n$-root of unity.
The traditional radix-2 Cooley-Tukey algorithm recursively decomposes the
computation between an odd and even part:
$$ \{X_k\} = \left(\sum_{j=0}^{(n-1)/2} x_j w_{n}^{k(2j)} + \sum_{j=0}^{(n-1)/2} x_{2j + 1} w_{n}^{k (2j + 1)} \right),~k\in[1, n]$$
This decomposition is called \emph{decimation in time} (\ditend).
An alternate decomposition performs \emph{decimation in frequency} (\difend):
$$ \{X_k\} = \left(\sum_{j=0}^{(n-1)/{2}} x_j w_{n}^{kj} + \sum_{j=(n-1)/{2}}^{n} x_{j} w_{n}^{kj} \right),~k\in[1, n]$$
When $n$ is a power of $2$, both decimations recursively decompose into a
perfectly balanced tree and take advantage of the symmetry properties of the
roots of unity.
The dataflow graph for the radix-2 \fft has a butterfly shape and is a good way of
visualizing the computations.
There is a symmetry between \dit and \dif in both the order of operations applied
and in whether the input or the output order is shuffled (Figure \ref{fig:butterflies}).

\begin{figure}
\centering
\begin{minipage}{.5\textwidth}
  \centering
  \includegraphics[width=\linewidth]{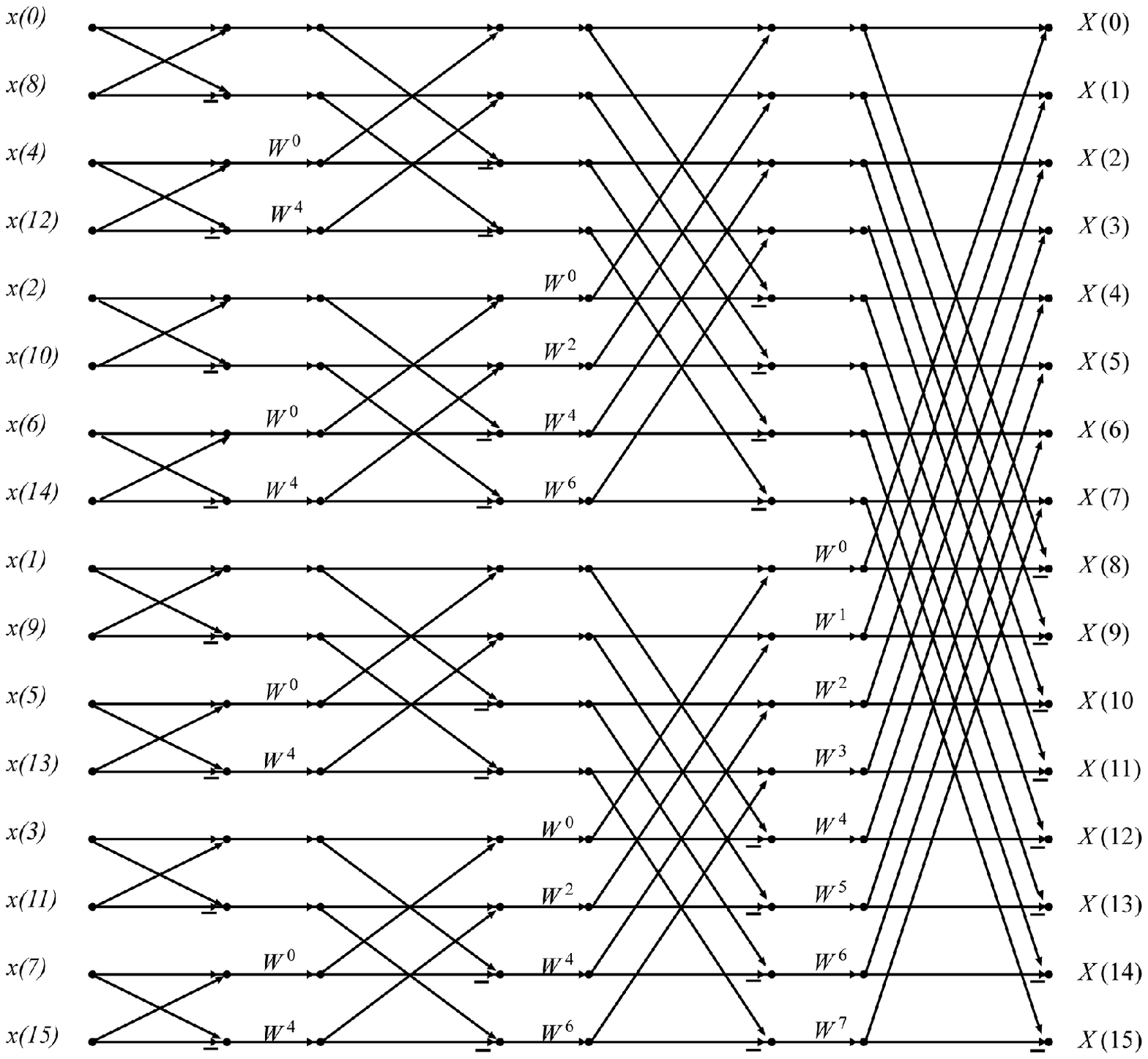}
\end{minipage}%
~~
\begin{minipage}{.5\textwidth}
  \centering
  \includegraphics[width=\linewidth]{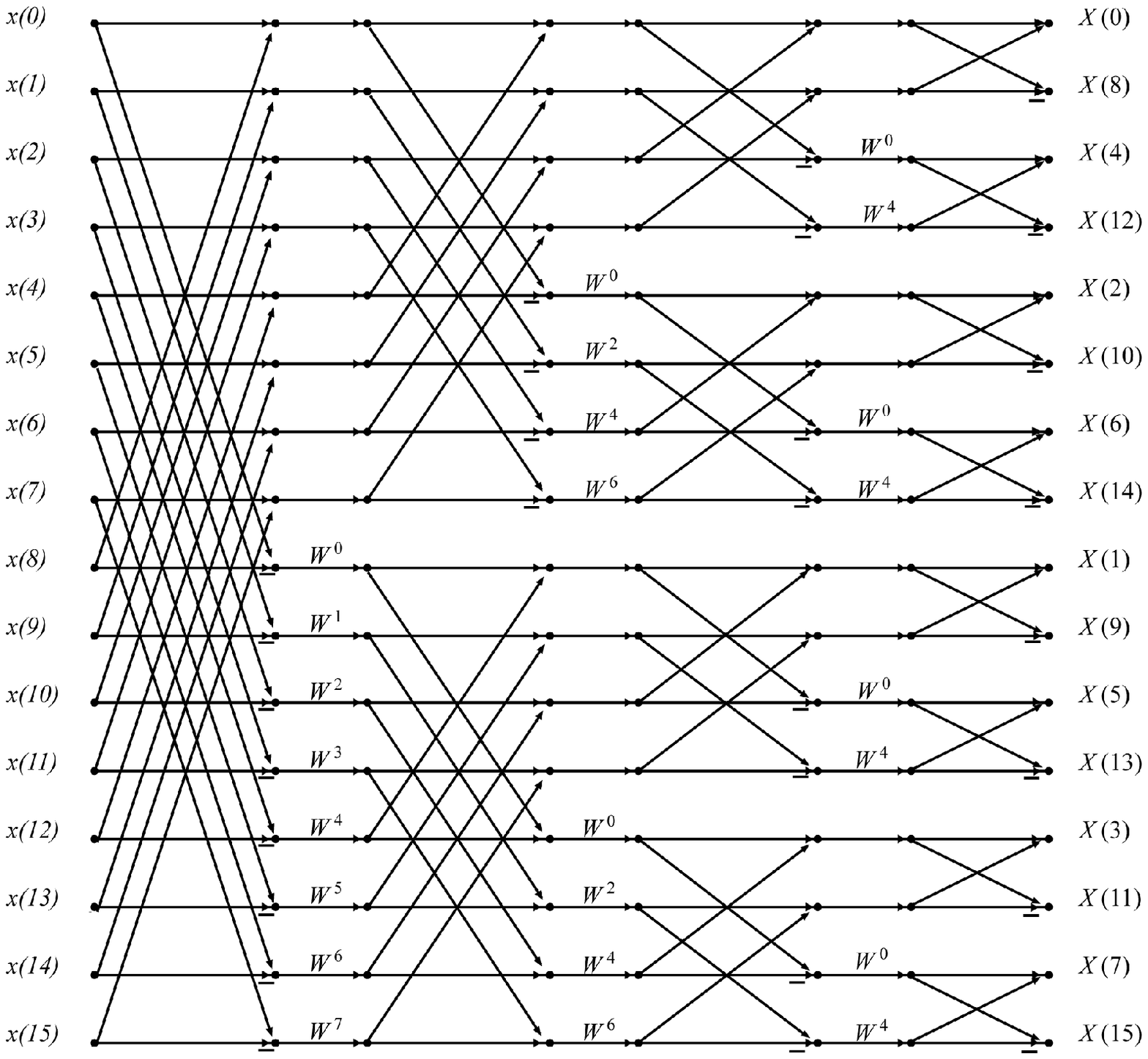}
\end{minipage}
\caption{\dit output ordered (left); \dif input ordered (right) (\cite{DITDIF})}\label{fig:butterflies}
\end{figure}

\subsection{\gpu Programming}
\label{sec:gpu}

There are a variety of references available
that describe \cuda and \nvidiaend's various \gpu
architectures (\cite{gpuoverview}) which we won't discuss in
detail, but the implementation of \fbfft very much depends upon specifics of
the Kepler \gpu architecture.

\nvidia \gpus execute code at the granularity of a \emph{warp} which is
defined as a set of $32$ threads in all existing architectures; each thread is assigned
a \emph{lane} within the warp.
These threads execute in a \emph{SIMT} (single instruction, multiple
thread) fashion, meaning that a warp is an atomic unit of execution.
It holds a single \emph{program counter} (PC) and can thus only execute a single
instruction at a time across all of its threads.
Collections of warps are brought together in \emph{blocks} or \emph{CTAs},
which together share a region of fast \emph{shared memory} resident on chip.
Blocks themselves can only exchange data via much slower \emph{global memory},
resident on the GPU or in the host CPU's address space.

Individual threads within a warp are free to take divergent paths, but since a
single PC is present, each branch in the execution will be serialized.
Threads that aren't participating in the branch in question are disabled.
In other words, if all $32$ threads were to take divergent code paths, we would
obtain only $1/32 \times$ of the computational efficiency.

Divergent code paths are hard to avoid, but the \nvidia instruction set has means to reduce their
cost (\cite{warpdivergence}). One is with predicated instructions, which are used for small branches,
in which all warp threads execute both parts of the branch, with non-participating threads having no
side effects.

Block threads have access to a register file, with up to $255$ registers per thread
for Kepler. Registers are allocated statically by the \cuda compiler.
An important performance factor when writing \cuda kernels is that data should
be kept in registers as much as possible to avoid communications. Registers in
\cuda are ``addressable'': it is possible to declare a static array within
registers and operate on its elements. The limitation is that all addressing
should be performed using statically determined constants so the compiler can
translate these accesses to a register number known at compile time. Indirect
addressing is also supported but results in copies to a local region within
global memory, which essentially constitutes register spilling.
Even with the presence of caches, using local memory usually comes with a
performance hit.\footnote{There are bleeding edge cases where a little local
  memory consumption helps performance; for instance, when restricting the
  number of registers per thread to increase occupancy.}
As a consequence, we design our kernels using aggressive inlining, template
parameters and unrolling directives to make all register accesses statically
addressable.

The Kepler architecture introduced specialized shuffle instructions
to exchange data between registers within a warp synchronously, which
avoids round-trips to shared or global memory.
Interestingly, these shuffle instructions allow the dynamic indexing of an
array held in registers, as long as the array is distributed in a cyclic
fashion across registers in each thread within a warp.

\begin{verbatim}
  float arr[3];
  ...
  // This simulates a linear array float realArr[96]:
  // arr[0] holds elements 0-31 (lane i holds element i)
  // arr[1] holds elements 32-63 (lane i holds element 32 + i)
  // arr[2] holds elements 64-95 (lane i holds element 64 + i)
  // Example: all warp threads read value held at realArr[34]
  float val = __shfl(arr[1], 2); // `1` must be statically known
                                 // `2` can be dynamic
\end{verbatim}

Many warps run in parallel and can be switched by the \gpu hardware at each
cycle. When enough parallelism is available (measured in \emph{occupancy} of the \gpu
as a first approximation), long latency operations are hidden
thanks to fast context switching.
Registers and shared memory come in finite quantities on each \gpu compute
multiprocessor. These limited resources are partitioned by the compiler and
the hardware amongst computations at the level of a \cuda
kernel.
Increased usage of registers or of shared memory can reduce \gpu \emph{occupancy},
which limits the ability to hide long latency operations. Reduced occupancy
does not necessarily result in performance loss (\cite{volkovoccupancy}).
There are often non-obvious performance tradeoffs in increasing or decreasing threads per
block, shared memory per block or registers per thread that are hard to
discover. This problem is one of the many reasons why designing a one-size-fits-all
implementation that aims to be efficient for any problem is difficult.

\end{document}